\documentclass[fleqn,10pt]{olplainarticle}

\usepackage{graphicx}
\usepackage{float}
\usepackage[]{subfig}
\usepackage{multirow}
\usepackage{amsmath}
\usepackage{verbatim}
\usepackage{pdfpages}
\usepackage{wrapfig}
\usepackage{colortbl}

\usepackage{flushend}

\title{Increased Complexity of a Human-Robot Collaborative Task May Increase the Need for a Socially Competent Robot}

\author[1,2]{Rebeka Kropivšek Leskovar}
\author[1,2]{Tadej Petrič}

\affil[1]{Department for Automation, Biocybernetics and Robotics,Jo\v{z}ef Stefan Institute, Jamova cesta 39, Ljubljana, Slovenia}
\affil[2]{Jo\v{z}ef Stefan International Postgraduate School, Jamova cesta 39, Ljubljana, Slovenia}


\begin{abstract}
An important factor in developing control models for human-robot collaboration is how acceptable they are to their human partners. One such method for creating acceptable control models is to attempt to mimic human-like behaviour in robots so that their actions appear more intuitive to humans. To investigate how task complexity affects human perception and acceptance of their robot partner, we propose a novel human-based robot control model for obstacle avoidance that can account for the leader-follower dynamics that normally occur in human collaboration. The performance and acceptance of the proposed control method were evaluated using an obstacle avoidance scenario in which we compared task performance between individual tasks and collaborative tasks with different leader-follower dynamics roles for the robotic partner. The evaluation results showed that the robot control method is able to replicate human behaviour to improve the overall task performance of the subject in collaboration. However, regarding the acceptance of the robotic partner, the participants' opinions were mixed. Compared to the results of a study with a similar control method developed for a less complex task, the new results show a lower acceptance of the proposed control model, even though the control method was adapted to the more complex task from a dynamic standpoint. This suggests that the complexity of the collaborative task at hand increases the need not only for a more complex control model but also a more socially competent control model.
\end{abstract}

\begin{document}

\flushbottom
\maketitle
\thispagestyle{empty}

\section{Introduction}\label{sec:intro}
As robots working in direct contact with people is becoming a more common practice, the question of effective human-robot collaboration (HRC) becomes more imperative each day. Due to this, a great amount of state-of-art research has been done on this subject over the past decade. One of the main issues being "\textit{How do we develop robots that will be accepted by their human partners?}" \cite{Brohl2019}. 

From an engineering standpoint, this manifests as HRC research mostly focusing on the development of robot control systems that are capable of replicating human behaviour in robots, as studies such as \cite{Noohi2016} showed that human-like behaviour in robots has a positive effect on human perception of their robot partner, as well as on the task performance. Here, they found that when collaborating with a robot, people find human-like behaviour more intuitive.  Furthermore, a study by Ivanova et al. \cite{Ivanova2020} found that humans can even prefer a robot partner with human-like behaviour to an actual human partner, which was further confirmed in our previous study \cite{LeskovarICAR2021}.

In this regard, many human-based control models had already been developed, such as \cite{Leica2016,Petric2017,Khoramshahi2018,Li2020}. However, the proposed robot control models in these studies only focus on imitating human behaviour \textit{in} a robot partner, disregarding the behaviour and dynamics \textit{between} the partners, which innately occur during human collaboration. These social interactions are a key component in any human collaboration experience and should be considered in HRC development as well. An example of social behaviour that naturally occurs in human collaboration is the leader-follower dynamics, where one of the partners takes on the role of a leader, while other people in the group follow, which allows the group to perform the task successfully. The importance of role distributions in HRC has been addressed in \cite{Jarrasse2014}. 

In our previous study \cite{Leskovar2021}, we tackled the question of leader-follower role allocation by first studying the dynamics in human dyads during a collaborative reaching task. We examined how the leader-follower dynamics occur during human collaboration and whether such roles are predetermined or not. The results provided in the study implied that the subjects who performed higher in the individual experiment would naturally assume the role of a leader when in physical collaboration with another person. From this, we developed a robot control method, described in \cite{LeskovarICAR2021}, that takes into account the leader-follower dynamic as they occur in human collaboration and allows the robot partner to assume both the role of a leader and that of a follower. However, the proposed control model was created only for the performance of a simple 2D reaching task. Here, the question of "\textit{what happens with the effectiveness and acceptance of such a robot control model when the collaborative task is more complex?"} remains.

In this study, we propose to extend upon our previous robot control method described in \cite{LeskovarICAR2021} by creating a novel robot control system through which a robot is capable of reaching a specific target on screen while avoiding obstacles just as a human would - by themselves or in collaboration with a human. This is done by first conducting a study on the performance of obstacle avoidance in human dyads, based on which a virtual Obstacle Avoidance Model is defined. Based on this virtual model, the robot control system is then developed, which is capable of creating human-like behaviour during obstacle avoidance, specifically curated to suit the human partner's natural behaviour. Furthermore, as in our previous paper, the leader-follower dynamics are implemented so that the robot can assume whichever role is preferable. The proposed system is then evaluated in a preliminary study by observing how human subjects respond to collaborating with a robot using the developed control system. This is done by evaluating both objective task performance as well as subjective task performance, using the Nasa Task Load Index assessment. Here, we hypothesise that our proposed control system will enable subjects to achieve better task performance during collaborative sessions compared to their individual performance. Furthermore, we hypothesise that the subject's subjective assessment of the collaborative task will be positive, as was the case in our previous study on HRC \cite{LeskovarICAR2021}.  

\section{Human collaboration study}\label{sec:humancollab}

\subsection{Subjects}
The study on human collaboration during obstacle avoidance included twelve male and four female participants, with all subjects having no prior experience with the experimental setup. Before the start of the experiment, the subjects were informed about the experimental procedure, potential risks, the aim of the study and gave their written informed consent in accordance with the code for ethical conduct in research at Jožef Stefan Institute (JSI). This study was approved by the National Medical Ethics Committee (No.: 0120-228/2020-3).

\subsection{Experimental Protocol}

\begin{figure}[t]
    \centering
    \includegraphics[clip,trim=0cm 0cm 0cm 0cm,width=0.75\textwidth]{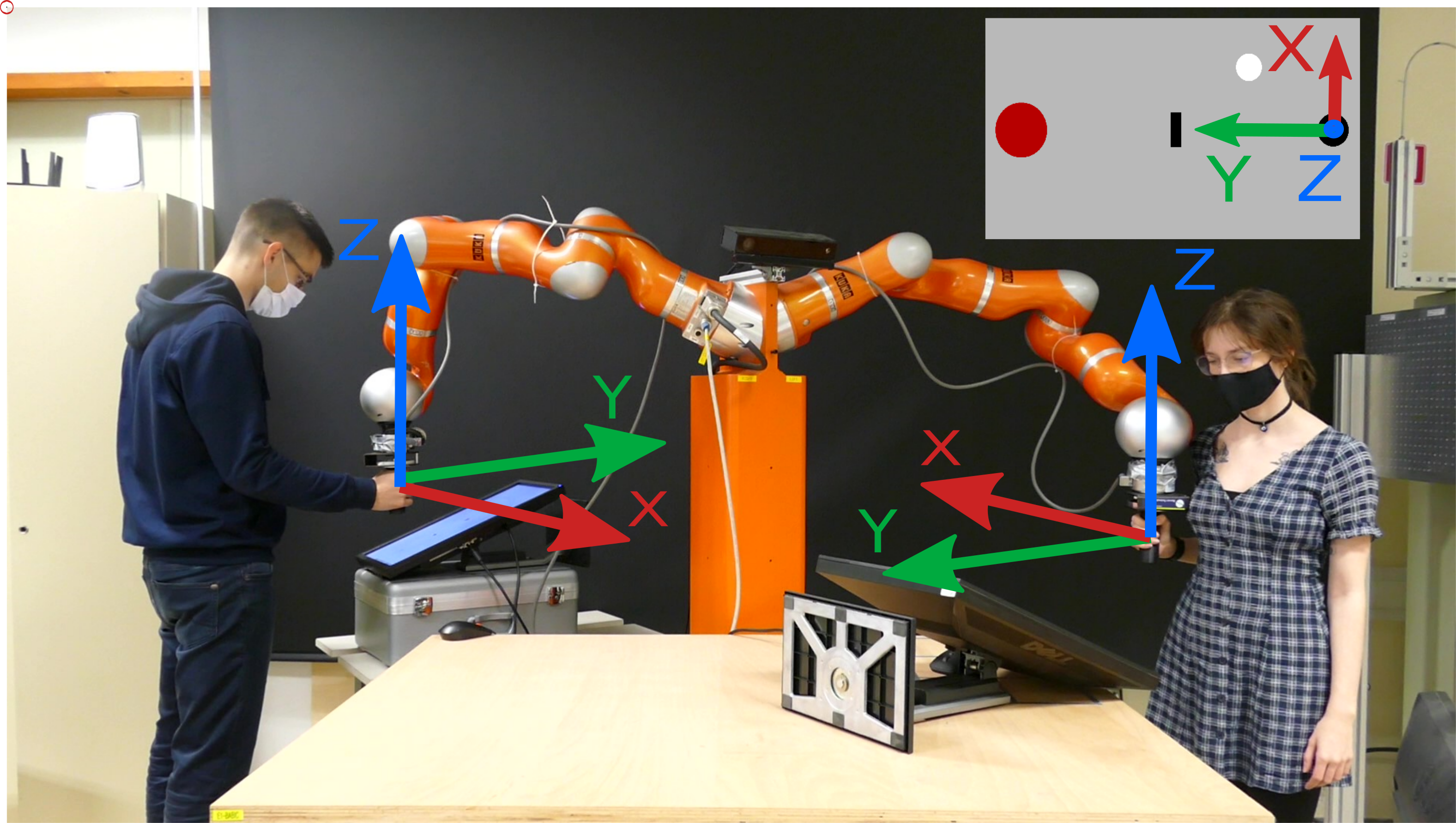}
	\caption{Picture of the experimental setup with two of the participants. The sketch above the right participant shows the graphic user interface that the subjects saw on the screen. It consisted of a starting point (black dot), target point (red dot), the obstacle (black line) and the controlled point (white dot) which the subjects moved either individually or together, when in collaboration.
	}\label{fig:experiment}
\vspace*{-0.4cm}
\end{figure}

Experiments on human collaboration were conducted on a dual-arm Kuka LWR robot, seen in Fig. \ref{fig:experiment}. The robots were used as a haptic interface between the subjects and the virtual environment, where they were moving a virtual point in the 2D environment shown and described in Fig. \ref{fig:experiment}. Here, the robot arms were used as two separate haptic interfaces for individual tasks or as one combined haptic interface, where the robot arms and their users were coupled together and controlled the same virtual point. This was achieved using a virtual dynamic model for the robot arms developed in \cite{Leskovar2020}. To match the 2D environment of the graphic user interface, the movement of the robot arms was limited to a 2D plane. This was done by constraining the z-axis of the robot's end-effector to a static position so that the angle between the subject's arm and forearm was 90deg in the starting position. 

The experiment consisted of two sets. In the first one, the subjects performed the required task individually with no obstacles present, while in the second set the subjects were coupled together to perform the task in collaboration. Moreover, in the second experimental set virtual obstacles were included in the task.

Throughout the experiment, the subjects were not told which task type they were performing. However, when they were coupled together, the two subjects could feel an external force produced by their partner. This established an open channel for haptic communication between the two subjects, which allowed them to sense when they were performing the task together or alone and their partner's action.

Each experimental set began with the subjects moving the controlled point on the screen to its starting position. When the controlled point was in its starting position a random target appeared on the screen. The subjects were instructed to reach this target and stay inside until the target disappeared. This was to prevent the subjects from simply running over the target without aiming for it. When the target disappeared, the subjects had to return to the starting position. The reaching task was repeated 90 times in each set in which 9 different targets with varying distances (5cm, 15cm and 25cm) and size (small, medium and large) were used in random order.

The collaborative experimental set was conducted in the same manner as the individual experimental set with the addition of obstacle avoidance. The obstacle appeared on the screen at the same time as the target and was always positioned midway between the starting position and the target. For instance, when the target distance was 5 cm, the obstacle appeared on the screen at a 2.5 cm distance from the starting position. If the subjects were unsuccessful in avoiding it, they failed the task and had to return to the starting position without reaching the target.

\subsection{Obstacle Avoidance Model}\label{sec:oamodel}

Based on the gathered data from the human experiments an obstacle avoidance model was created. Here, our goal was to create a virtual model that is capable of generating the same movement as an individual that is performing the same task.

In the analysis of the data gathered during human collaboration, we have found that subjects performed two separate tasks during the experiment. The first one being the task of reaching the target and the second one being the task of avoiding the obstacle. We gathered this by observing the force trajectories of both subjects during the reaching task where obstacles were present. Here, we could see that subjects who saw the obstacle were primarily focused on avoiding it by exerting all of their force in the direction away from the obstacle. After successfully avoiding it, however, the subjects moved all their force and focus towards reaching the target in a straight line.

Due to this, we constructed the virtual obstacle avoidance model as such - by combining the description of the movement towards the target using Dynamic Movement Primitives (DMPs) with a potential force field around the obstacle. 

This was done by first describing the movement of the virtual point similar as a linear Dynamic Movement Primitive (DMP), which is given by:
\begin{equation}\label{eq:lindmp}
    \tau\dot{v} = K(g-x) - Dv,
\end{equation}
\begin{equation}
    \tau\dot{x} = v,
\end{equation}
where $x$ and $v$ are position and  velocity of the system, and $g$ the position of the target or goal. $K$ corresponds to the spring constant and $D$ to the damping constant of the system, which were adapted based on the person's natural movement.

Here, we found that both $K$ and $D$ are not constant per se, but rather change based on the type of target displayed on the screen. More specifically, the constants $K$ and $D$ depend on the Index of Difficulty ($ID$) of the target, described in \cite{Fitts1954}. As such, the constants are defined as:
\begin{equation}
    K = k_1ID + k_2
\end{equation}
\begin{equation}
    D = k_3ID,
\end{equation}
where parameters $K_1$ to $k_3$ are determined based on the measured movement of a person performing the same task.

In order for the model to successfully avoid obstacles, a potential field was added to the linear DMP defined in Eq. \ref{eq:lindmp}. The combination of the linear DMP and a potential field is based on the work of Park et al. in \cite{Park2008} and is described as:
\begin{equation}\label{eq:lindmpobs}
    \tau\dot{v} = K(g-x) - Dv + \varphi(x,v),
\end{equation}
where $\varphi(x,v)$ is the repellent acceleration force used and is defined as a negative gradient of the dynamic potential field around the obstacle.

This is described in the work of Park et al. \cite{Park2008} as:
\begin{equation}\label{eq:cdmp}
    \varphi(x,v) = \lambda(-cos\theta)^{\beta-1} \frac{\|v\|}{p}\left( \beta\nabla_x cos\theta - \frac{cos\theta}{p}\nabla_x p\right),
\end{equation}
where $p$ is the distance between the current position and the obstacle and $\theta$ the angle between the current velocity direction and the direction towards the obstacle. $\lambda$ is a constant that modifies the strength of the entire field and $\beta$ is an additional constant for correction, presumably.

Using the gathered data from human experiments we tried to learn how the constants $\beta$ and $\lambda$ change based on the task. We have found that the values of both $\beta$ and $\lambda$ change according to the distance between the starting point and the obstacle. These were best described as:
\begin{equation}
    \lambda = k_1/o  + k_2
\end{equation}
\begin{equation}
    \beta = k_3o^2 + k_4o + k_5,
\end{equation}
where $o$ is the distance between the starting point and the obstacle and $k_1$ through $k_5$ are parameters which are specific to each individual subject. These parameters were determined using the optimisation procedures.

\begin{figure*}[]
    \begin{center}
        \includegraphics[clip,trim=1.8cm 0.5cm 1.8cm 0cm,width=\textwidth]{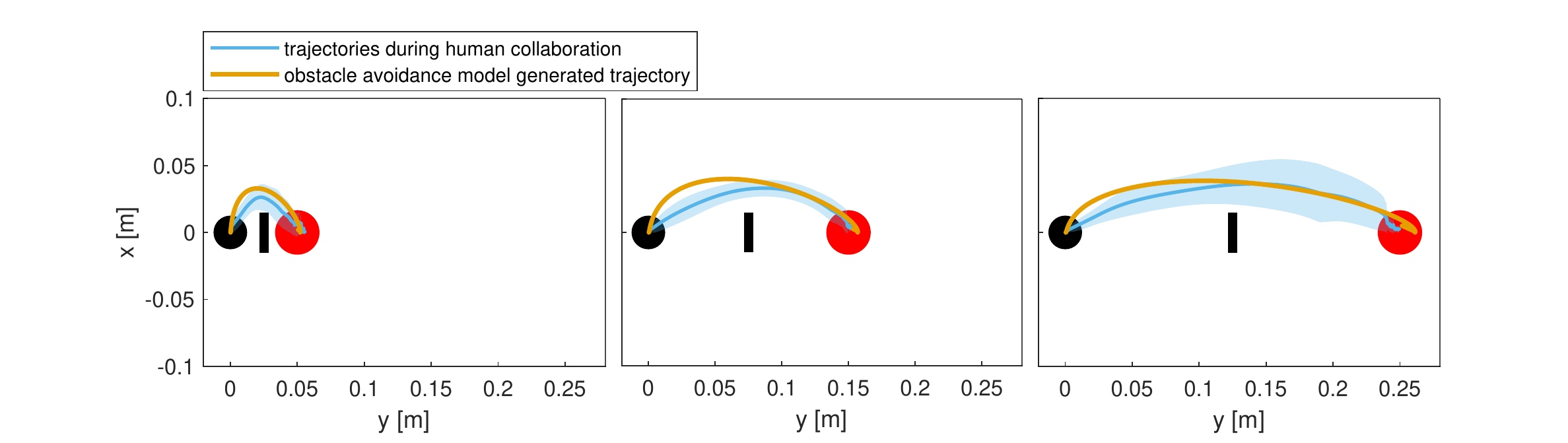}
        \caption{Sketch of average movement of the virtual point during human collaboration experiments and the generated robot trajectory based on the Obstacle Avoidance Model. The sketch showcases the movement of a typical pair for all target distances used in the experiment.} \label{fig:motion}
    \end{center}
\end{figure*}

\section{Human-robot collaboration study}\label{sec:hrc}
Based on the obstacle avoidance model found through the human collaboration study, a robot control model was developed to investigate whether such a model can be successfully implemented into human-robot collaboration as well.

\subsection{Control method}
The robot control system used in this study was based on the virtual dynamic model which is the same in the human collaboration study for haptic communication. This ensured that the differences between human collaboration and HRC were kept to a minimum.

As described in our previous paper \cite{Leskovar2020}, the virtual dynamic model used in our study controls the movement of the virtual point based on the amount of force applied to the end-effector of the robot. Therefore, our goal was to create a virtual control model that was able to generate a force trajectory that would allow a robot to perform the same task of obstacle avoidance as seen in the human experiment individually.

Here the virtual dynamic model is used to control a virtual point individually or in collaboration, by coupling the two human subjects together. To enable the virtual model for human-robot collaboration the force equation, described in \cite{Leskovar2020} has been redefined as:
\begin{equation}
    \boldsymbol{F} = \boldsymbol{F_r} + \boldsymbol{F_h},
\end{equation}
where $F_r$ becomes the force applied by a virtual robot partner and $F_h$ becomes the force applied by the human to the end-effector. Note that the virtual robot partner is able to perform the task alone, just as a human partner could.

In the proposed control system the force of the robot $F_r$ is based on the Obstacle Avoidance model described in the previous section \ref{sec:oamodel} and is described as:
\begin{equation}\label{eq:fr}
    \boldsymbol{F_r} = K_l\cdot(K_p\cdot(\boldsymbol{p} - \boldsymbol{p_t}) + K_d\cdot(\boldsymbol{v} - \boldsymbol{v_t}),),
\end{equation}
where $K_p$ and $K_d$ are the spring and dampening coefficients of the robot control system, $\boldsymbol{p}$ is the current position of the controlled point while $\boldsymbol{p_{t}}$ is the desired position of the controlled point, based on the desired movement generated by the Obstacle Avoidance Model, and lastly $\boldsymbol{v}$ is the current velocity of the controlled point, while $\boldsymbol{v_{t}}$ is the desired velocity, which was also based on the desired movement generated by the Obstacle Avoidance Model.

The desired movement and velocity of the controlled point depended on the human partners natural performance of the same task. This was determined based on the measurements taken during an individual performance of the same obstacle avoidance task.

$K_l$ in Eq. \ref{eq:fr} presents the coefficient that defines the leader-follower dynamics in HRC, which was previously studied in \cite{Leskovar2021} and is defined as:
\begin{equation}
    \Delta K_l  
    \begin{cases}
       < 1  &\mbox{;} \textrm{ robot follows the human} \\
       = 1 &\mbox{;} \textrm{ robot is equal to human} \\
       > 1 &\mbox{;} \textrm{ robot leads the human} \\
    \end{cases}
\end{equation}

In the current preliminary study, the leader coefficient was the same for all subjects and was predetermined as $K_l = 0.75$ for cases where the robot followed the human, $K_l = 1$ for cases where robot and human were equal partners, and $K_l = 1.25$ for cases where robot immitated the role of a leader within the collaborative dyad. 

Furthermore, as the parameters of the Obstacle Avoidance Model change based on the position and size of the target as well as the position of the obstacle, the robot partner needed to be capable of discerning what the goal of each trial was in real time. This was done by implementing real-time tracking of the target's position and size into the robot control system, upon which the system was able to adjust its force trajectory accordingly.

\subsection{Evaluation of the robot control system}
In order to test the proposed control model, a preliminary study was performed, with 4 subjects participating. Here, two of the subjects were female and two were male.

The proposed robot control system was evaluated using the same setup, seen in Fig. \ref{fig:experiment}, and the same reaching task, described in Section \ref{sec:humancollab}. Here the experiment consisted of four different sets, where subjects performed the reaching task individually or in collaboration with a robot partner that leads, a robot partner that follows and a robot partner who is their equal. Note that in HRC the robot partner was a virtual partner rather than a physical one, meaning the forces applied by the robot partner manifested in physical form as an autonomous movement of the robot arm the subject was holding.   

Each experiment set consisted of 45 trials, accumulating to 180 trials altogether. To mitigate the influence of learning on subjects' performance in different experiment sets, the order in which experiment sets were performed was different for each subject, with the individual set always being the first one as the data from this trial was used to generate the robot partner's behaviour.

After each experiment set the subjects had a 5-minute break during which they were asked to fill out the Nasa-TLX form, which was used to assess the subject's subjective workload during the experiment. 

The evaluation of the proposed robot control method further included an objective analysis of task performance during each experiment set. The task performance was defined using Fitts' law's index of performance (IP) \cite{Fitts1954}, which is described as:
\begin{equation}\label{eq:ip}
    IP = \frac{ID}{MT} ,
\end{equation}
where $MT$ is the measured movement time and $ID$ is the index of difficulty,
which has several formats in literature as seen in \cite{Fitts1954},\cite{Zhai2004}. In this study the Shannon formulation \cite{MacKenzie1992} was used, which is defined as:
\begin{equation}\label{eq:id}
    ID = log_2(\frac{D}{W} + 1) .
\end{equation}

\subsection{Results}

\begin{figure}[!b]
    \begin{center}
        \includegraphics[clip,trim=0cm 0cm 0.5cm 0.3cm,width=0.75\textwidth]{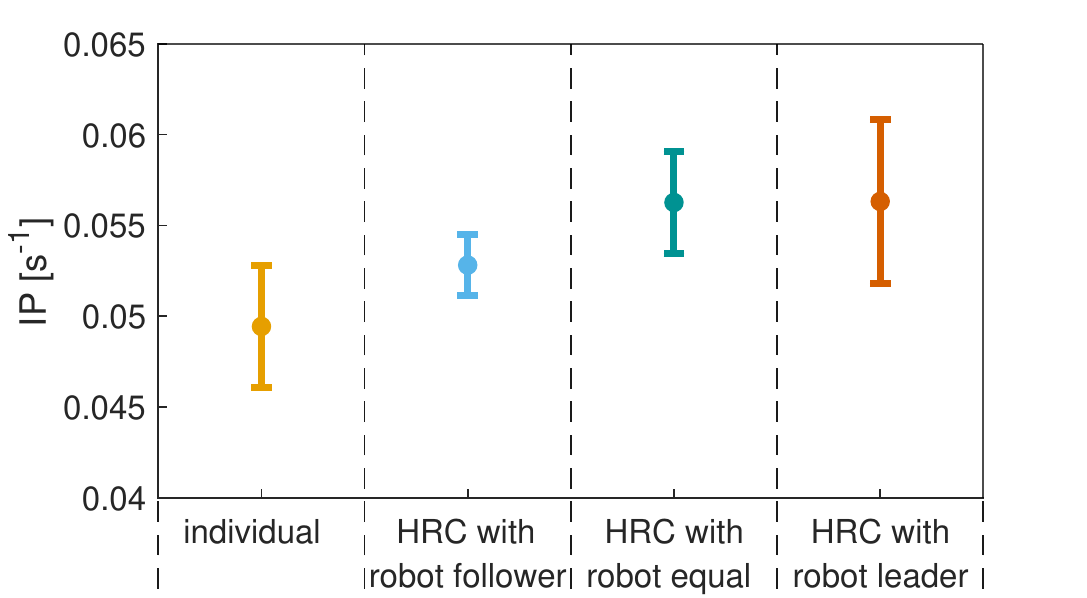}
        \caption{Average Fitts' law's Index of Performance (IP) for each experiment set, calculated as described in \cite{Fitts1954}. The higher the IP of an experimental set, the better the performance of a task was.} \label{fig:ip}
    \end{center}
\end{figure}

Looking at Fitts Law's Index of Performance in Fig. \ref{fig:ip}, we see that all robot collaboration improved objective performance compared to the individual performance of a person. This result correlates with our previous study on human-robot collaboration \cite{Leskovar2021}, where both human and robot partners improved the overall task performance.

\begin{table}[h]
\caption{Number of collisions for each subject in each experiment. The number of collisions represent the total amount of collision throughout the experiment set, for all targets.}
\centering
\begin{tabular}{|l|c|c|c|c|}
\hline
\multicolumn{1}{|c|}{\cellcolor[HTML]{C0C0C0}\textbf{experiment sets}} & \multicolumn{1}{l|}{\cellcolor[HTML]{C0C0C0}\textbf{subject 1}} & \multicolumn{1}{l|}{\cellcolor[HTML]{C0C0C0}\textbf{subject 2}} & \cellcolor[HTML]{C0C0C0}\textbf{subject 3} & \multicolumn{1}{l|}{\cellcolor[HTML]{C0C0C0}\textbf{subject 4}} \\ \hline
\textbf{individual task} & 0/45 & 0/45 & 0/45 & 0/45 \\ \hline
\textbf{robot follower collaboration} & 1/45 & 5/45 & 1/45 & 2/45 \\ \hline
\textbf{robot equal collaboration} & 4/45 & 4/45 & 2/45 & 2/45 \\ \hline
\textbf{robot leader collaboration} & 5/45 & 4/45 & 2/45 & 4/45 \\ \hline
\end{tabular}\label{tab:collisions}
\end{table}

However in Table \ref{tab:collisions}, we can see that the number of collisions increase when subjects are in collaboration with a robot. Furthermore, it should be noted that most of the collisions occurred in cases where the obstacle was closest to the starting position. 

\begin{figure*}[!b]
    \begin{center}
        \includegraphics[clip,trim=3.5cm 0.5cm 3cm 0cm,width=\textwidth]{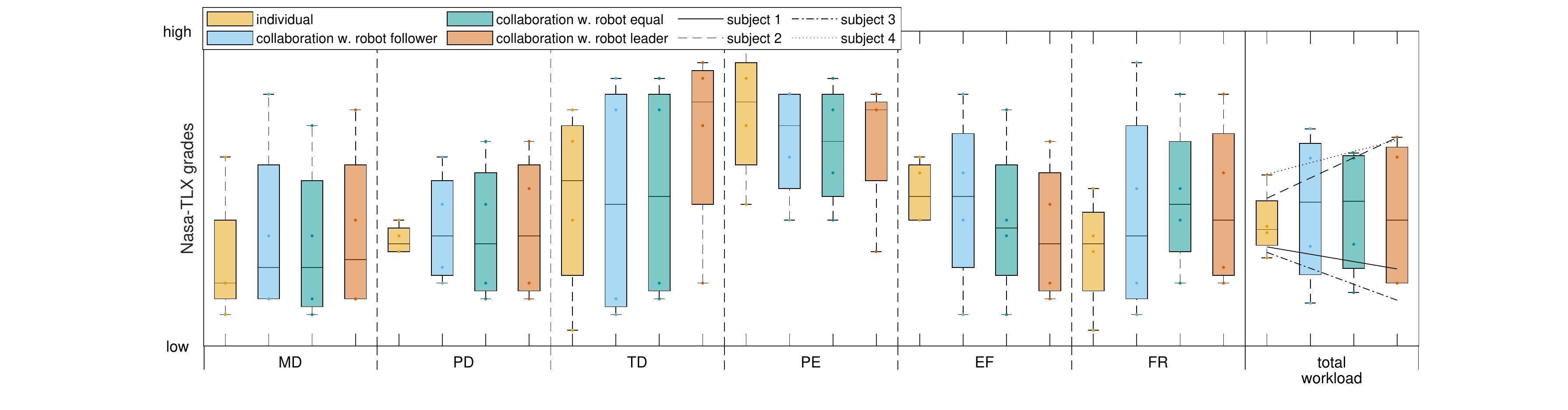}
        \caption{Nasa-TLX scores for all experimental sets, where each factor is shown separately. Here, MD stands for mental demand, PD for physical demand, TD for temporal demand, PE for performance, EF for effort and FR for frustration. The graph also shows the total Nasa-TLX score for each experimental set. Here the weights for each factor based on individual subject's preference were taking into account when calculating the total workload.} \label{fig:tlx}
    \end{center}
\end{figure*}

Moving our focus to subjective performance evaluation, based on results from the NASA-TLX questionnaire, we can see two distinct reactions to the human-robot collaboration tasks. In Fig. \ref{fig:tlx} we can see that for 2 of the 4 subjects the overall task load increases with the presence of the robot, while for the other 2 subjects the workload seems to decrease with the increased influence of the robot.

\section{Discussion}\label{sec:discussion}
Results of the Fitts' Law's Index of Performance shown in Fig. \ref{fig:ip} indicate that overall performance did increase in all cases where subjects were collaborating with a robot. This confirms the findings from our previous paper \cite{LeskovarICAR2021}, upon which the current control system is based on. The results from this study state that task performance increases when subjects collaborate with either another human or a robot, as the additional forces increase the speed at which the task can be accomplished.

However, unlike in our previous study, the collaborating dyad had an additional task of avoiding an obstacle in order to reach a target or face failure. This means that only the successful trials were able to be taken into account when calculating Fitts' Law metrics. Due to this, the number of failed trials, i.e. the number of collisions occurring in each experimental set, should be looked at as well to determine how successful the proposed model is at collaborative obstacle avoidance. Here, we can see from Table \ref{tab:collisions} that collisions increase when subjects were collaborating from 0 collisions, up to 5 collisions out of 45 trials, meaning that in regards to the robot helping the human avoid an obstacle, our proposed model did not perform well. A reason for this could be a lack of communication between the robot and the human partner that would allow them to actively coordinate their movement to avoid the obstacle. 

As stated in \cite{Cohen1991}, communication is an integral part of successfully performing any collaborative task. For instance, when looking at the human collaboration the same problem seemed to occur, with one of the human partners hindering the other's ability to avoid the perceived obstacle by pushing them in a different direction. We can presume that this is due to the two partners' plans of action differing from each other and there being no effective way for them to coordinate with each other in real-time. Although they felt each other's force through the haptic channel of communication, this seemed not to be enough when neither of the individuals wanted to conform to their partner's behaviour. Or in some cases, they might not have been able to course-correct fast enough, which can be confirmed through the fact that both in human collaboration and in HRC the highest rate of collisions occurred in cases where the obstacle was the closest to the starting point. 

Unlike in the human collaborative experiment, however, the lack of communication about their individual plan of action between the robot and the human partner could be easily fixed by displaying the planned movement of the robot on screen. This would allow for the human subject to understand the robot partner's actions better, which might further improve their sense of frustration during the collaborative task, observed in Fig. \ref{fig:tlx} for the current experiment. However, such adaptations can only be implemented from the perspective of the person collaborating with a robot, while the robot still has no understanding of their human partner's plan of action. This in turn decreases the amount of autonomy and adaptability of the robot, which in turn changes the leader-follower dynamics within the human-robot dyad. As the human is the only partner capable of making a fully informed decision, the robot partner is then automatically pushed into the follower role, while their human partner assumes the role of a leader.

The lack of clear communication and its detriments can be seen in the subjective assessment of the proposed robot control model as well. Looking at Fig. \ref{fig:tlx}, we can see an obvious rift between the opinions of the participants, where half of the participants felt an increase in the workload with the addition of a robot partner and half of the participants felt a noticeable decrease in the overall workload. Here, it is interesting to note that the subjects who felt an increased workload during HRC also rated higher workload in the individual tasks compared to the subjects who sensed a decrease in the overall task workload when in collaboration. This suggests that there may be a dissonance between the subjects' perception of the task as a whole, pointing to some psychological aspects coming into play such as a person's individuality, their task execution preference, need for control etc. However, before making any further claims, another evaluation needs to be done with a larger and broader pool of participants in order to gain a clearer picture of how the proposed robot control method is perceived.

\section{Conclusion}
In this study, we describe a proposed robot control model that is capable of avoiding obstacles autonomously as humans do. In addition, it is able to perform the same task in cooperation with a human partner, where it can take into account the leader-follower dynamics observed in human cooperation by performing the role of leader, follower, or equal. Furthermore, the aim of the study was to create a robot control system that can be personalised based on the needs of the individual collaborating with the robot, thus not hindering a person's performance of the task.

The results of the HRC using the proposed control method show that it improves overall task performance compared to individual human performance. However, the proposed robot control method did increase the number of collisions during collaborative task execution and was not positively accepted by all participants.

Although the results from the preliminary study showcase lower acceptance of the proposed robot control model than hypothesised, the study highlights important factors which need to be considered in future development of HRC control models. As stated in the introduction, collaboration between people is much more than just a physical activity, it is also a social activity. Therefore, when trying to develop an acceptable robot control model for HRC the same, or even additional, social dynamics need to be taken into account.

The results provided in this study suggest that as the complexity of the collaborative task at hand increases, so does the complexity of human perception, communication patterns, and overall social interactions, which will need to be addressed accordingly if we want to develop acceptable control models for HRC in the future. Thus, we must strive to develop robots that are more socially competent, even if the task itself does not appear to be social but rather physical, such as transporting objects.

\section*{Declarations}

\subsection*{Funding}
This work was supported by Slovenian Research Agency grant N2-0130.

\subsection*{Conflict of interest}
The authors have no relevant financial or non-financial interests to disclose.

\subsection*{Ethics approval} 
This study was conducted in accordance with the code for ethical conduct in research at Jožef Stefan Institute (JSI) and was approved by the National Medical Ethics Committee (No.: 0120-228/2020-3, approved on 13.7.2020).

\subsection*{Consent to participate}
Prior to conducting the experiment, all participants were informed about the experimental procedure, potential risks, the aim of the study and gave their written informed consent in accordance with the code for ethical conduct in research at Jožef Stefan Institute (JSI).

\subsection*{Consent for publication}
All authors have read and agreed to the publication of the final version of the manuscript.

\subsection*{Availability of data, materials and code}
All materials, data and code used in this study are available upon request to the authors.

\bibliography{main}

\end{document}